# A Systematic Mapping Study on Testing of Machine Learning Programs

Salman Sherin[1,2], Muhammad Uzair khan[1,2], Muhammad Zohaib Iqbal[1,2]
[1] Quest Lab, National University of Computer and Emerging Sciences, Islamabad, Pakistan
[2] UAV Dependability Lab, National Center of Robotics and Automation, Pakistan
{salman.sheirn, uzair.khan, zohaib.iqbal}@questlab.pk

**Abstract**
**Context:** Machine learning (ML) has made tremendous progress in the last few years leading to usage in mission-critical and safety-critical systems. This has led researchers to focus on the techniques for testing ML-enabled systems, and has been further emphasized by recent hazardous incidents (e.g., Tesla car accident).

**Objective:** We aim to conduct a systematic mapping in the area of testing ML programs. We identify, analyze and classify the existing literature to provide an overview of the area.

**Methodology:** We followed well-established guidelines of systematic mapping to develop a systematic protocol to identify and review the existing literature. We formulate three sets of research questions, define inclusion and exclusion criteria and systematically identify themes for the classification of existing techniques. We also report the quality of the published works using established assessment criteria.

**Results:** we finally selected 37 papers out of 1654 based on our selection criteria up to January 2019. We analyze trends such as contribution facet, research facet, test approach, type of ML and the kind of testing with several other attributes. We also discuss the empirical evidence and reporting quality of selected papers. The data from the study is made publicly available for other researchers and practitioners.

**Conclusion:** We present an overview of the area by answering several research questions. The area is growing rapidly, however, there is lack of enough empirical evidence to compare and assess the effectiveness of the techniques. More publicly available tools are required for use of practitioners and researchers. Further attention is needed on non-functional testing and testing of ML programs using reinforcement learning. We believe that this study can help researchers and practitioners to obtain an overview of the area and identify several sub-areas where more research is required.

## 1. Introduction

Machine learning (ML) has made tremendous progress in the past few years by achieving human level performance in multiple tasks such as speech recognition, image classification and playing games like Alpha Go [1-3]. These advances have led to the adoption of machine learning in mission critical and safety critical systems like autonomous cars, malware detection, Unmanned Aerial Vehicles (UAV), fraud detection, heart failure detection and aircrafts collision avoidance systems. However, recent hazardous accidents related to such systems is a big concern. For example, Google self-driving car recently crashed into a bus due to unexpected conditions [4] and Tesla car hit a trailer without recognizing it as an obstacle [5]. To alleviate such accidents, there is a growing interest in researching testing techniques for ML systems.

However, the testing of ML systems poses new challenges for software testing community. First, these systems lack an explicit oracle. Second, these programs do not have control flow like traditional programs and cannot be tested with traditional software testing techniques which rely on explicit oracle and program control flow. Moreover, the functionality of ML systems depends on the set of data input to them; a small change in the training data can have significant influence on the behavior of the system and the results of the learning process. Such systems have been classified as "non-testable" by traditional methods by some researchers leading to the requirement of developing novel strategies to test them [6]. Several techniques are recently proposed in the literature for testing of ML programs [S1-S37]. These techniques can be used to analyze and test the generality and robustness of ML programs. However, there is no study that integrates and summarize these techniques to assist new researchers and practitioners in the field. As a growing area of research, we feel it is important to identify, analyze and classify the existing literature and provide an overview of the trends in this particular area. In this paper, we conduct a Systematic Mapping (SM) in the area of testing ML systems.

An SM is a type of systematic review used to review, classify and structure papers relevant to a specific research area in software engineering [7]. The aim of SM is to obtain an overview of the area by classifying and summarizing the existing literature. Unlike SLR, SM focuses more on the classification scheme and pose broader research questions [8]. SM studies are conducted in several other domains in software engineering, which helps in the assessment and efficient interpretation of the available knowledge [8]. The identification of trends and significant gaps provide a strong basis for future research in the area. The outcomes of SM can be a valuable asset for new researchers (e.g., PhD students) to identify potential gaps in the areas where research is lacking [9]. It helps practitioners who want to stay up to date with state of research and identify suitable techniques and tools.

In this paper, we conduct a SM by including the published papers related to the testing of ML systems. To the best of our knowledge there is no systematic review that systematically categorize and synthesize the published literature in this specific field of testing ML applications. The following are the main contributions of this study.
- The classification scheme for categorization of published papers on the basis of contribution facets, research facets and technique type
- The identification characterization of existing techniques and tools for testing of ML applications
- An analysis of empirical evaluations reported in the papers
- Quality assessment of the included studies
- An online repository for gathering and analyzing papers on testing of ML applications

The rest of the paper is organized as follows. Section 2 presents the summary of the related work. Section 3discusses the review protocol and search process to perform this study. Section 4 provides the results and discussion. Section 5 discusses threats to validity. Finally, section 6 concludes the study.

## 2. Related Work
In this section, we discussed several identified secondary studies related to our work. In general, four studies are identified as surveys and two studies are systematic reviews with one being a systematic mapping and another being a systematic literature review. Surveys are typically not conducted in systematic and unbiased fashion, and thus are considered by some researchers to have little scientific value [8, 9]. Also, they often suffer from selection bias and their results are not repeatable due to lack of an explicitly defined protocol and search strategy

[10]. In contrast, systematic reviews follow a well-defined protocol and search strategy which reduces the selection bias and increases the repeatability and reliability of results.

Kanewala *et al.* [11] presented a survey of techniques used to test scientific software having hard of Oracle. Scientific software is defined as the software used for the scientific purpose that require specialized domain knowledge and has higher complexity [12]. For example, an automated testing tool for MATLAB or software's which are developed for publishing a paper [13]. The paper focuses on three types of techniques for testing scientific software i.e., metamorphic testing, run-time assertions and oracles built with using ML techniques. However, it does not focus on analyzing techniques for testing ML programs in particular. Though ML programs can also be classified as programs without an oracle, however, the survey does not include any details regarding the techniques for testing ML programs. One of the reason might be its publication year, the survey is published in 2013 whereas most of the literature in this area (testing ML program) is available afterward. Our study presents a systematic mapping with an exclusive focus on the identification and classification of testing techniques for ML programs. We present a detailed analysis of recently proposed techniques to test ML programs. In addition, we also study attributes related to empirical evaluations and reporting quality of papers.

In another paper, Kanewala *et al.* [14] conducted an SLR by including papers related to testing scientific software. The SLR poses four questions regarding the definitions, faults, methods and challenges of testing scientific software. However, it does not focus on ML programs and have different goals, scope and research questions than ours.

Metamorphic testing (MT) is one of the prominent techniques for testing of programs having no oracle (including ML programs) because it does not require any oracle for implementation. However, it is also used as complementary technique with other testing strategies for conventional software testing having an explicit oracle. The main challenge in MT is the generation of metamorphic relations and is currently a growing area of research. Chen *et al.* [16] conducted a comprehensive survey to analyze and summarize literature on MT and discusses possible research gaps and challenges in the area of MT. The results of the study are complementary to our study as MT is the most popular technique for testing of ML programs.

The oracle problem in the area of software testing is extensively studied by different researchers and several approaches are proposed in the literature to provide solution to this problem. Barr *et al.* [15] conducted a comprehensive survey on oracle problem and discusses various target approaches available in the literature. The study is related to our study because it provides discussion on the construction of an oracle by using metamorphic testing which is one of the technique used to test ML programs. However, the primary focus of our study is to summarize and integrate the existing literature in the area of testing ML programs and rather than just metamorphic testing. Furthermore, we formulate different goals and pose a unique set of questions in a specific area to provide analysis of the published work in the area. We compare our results regarding metamorphic testing with this survey where possible.

Masuda *et al.* [17] conducted a survey that discusses the software quality methods for ML applications including software testing techniques. However, the survey failed to provide a thorough classification of techniques for testing ML programs that aims to position the existing literature for the identification of research gaps and future directions. Moreover, it does not follow a systematic protocol for identification and classification of papers like a systematic review. Therefore, the results are neither repeatable nor reliable due to the random selection and analysis of papers.

Patel et al. [6] presented a systematic mapping study of non-testable systems by discussing techniques used for testing of non-deterministic and stochastic systems in general. The article has no formal classification for mapping of the existing techniques and literature. In contrary, our article targets mainly the identification and classification of techniques used for testing of ML programs in particular. We do not discuss the testing of all type of non-deterministic systems but ML programs. We provide the classification of existing techniques and trends specifically for researchers working in the area of testing ML programs. Additionally, we also provide a quality assessment of the included papers. Our classification is based on a unique set of papers related to the testing ML programs and may not be suitable for testing of all non-deterministic systems. Furthermore, our paper poses a different set of research questions and has unique findings.

In this paper, we present a systematic mapping in the area of testing ML programs. We followed a well-established guideline of systematic mapping to identify and review papers. We identify themes and provide different classifications in the area. Moreover, we analyze the existing empirical evaluations and assess the reporting quality of all the included papers. Finally, we provide a detailed discussion on research gaps and future recommendations. We have discussed the uniqueness of our study in relation with all the other related studies, given in Table 1. However, some of our findings might be similar with the existing related studies. We believe that this similarity will strengthen our discussion and will reinforce the findings of the existing studies.

Table 1. An overview of the related studies

| Year | Title | Type | Focus |
| --- | --- | --- | --- |
| 2018 | A mapping study on testing non-testable systems | SM | Conducting systematic mapping study of non-testable systems |
| 2018 | A Survey of Software Quality for Machine Learning Applications | Survey | Survey on generic software quality methods for ML applications |
| 2017 | Metamorphic Testing: A Review of Challenges and Opportunities | Survey | Identifying the challenges in the application of metamorphic testing |
| 2015 | The Oracle Problem in Software Testing: A Survey | Survey | Identifying challenges of finding oracle and its remedies |
| 2014 | Testing Scientific Software: A Systematic Literature Review | SLR | Analyzing testing techniques for applications having complex or no oracle. |
| 2013 | Techniques for Testing Scientific Programs Without an Oracle | Survey | Identifying testing techniques for scientific programs having no oracle |

## 3. Methodology

This study is conducted by following the guidelines of systematic mapping study proposed by Peterson et al. [8] and Kitchenham et al. [9]. A systematic mapping study aims to identify and classify the studies published in the area. Therefore, after defining our research questions, query strings and search process we define our inclusion and exclusion criteria to carefully select studies for data extraction. The selected studies are then analyzed for the purpose of classification and findings regarding the trends and gaps in the area. Figure 1 shows an overview of our methodology discussed in detail in the subsequent sub-sections.

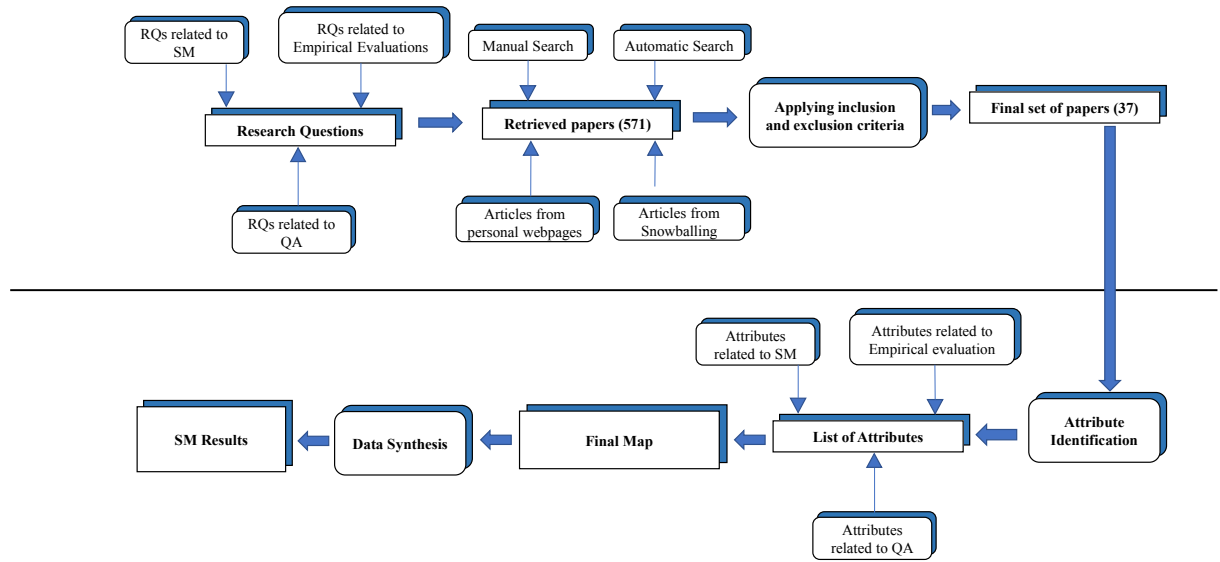

Figure 1. An overview of the protocol

## 3.1. Goals and Research questions

The goal of this study is to identify, analyze and synthesize the work published in the area of testing machine learning and deep learning systems. We aim to (1) systematically review the existing papers in the area to conduct the systematic mapping, (2) analyze the empirical evaluations and (3) assess the reporting quality.

Based on our goals, we formulate our research questions in four categories, i.e., systematic mapping, empirical evaluations and quality assessment. To extract detailed information, each category consists of specific research questions, as described below.

### 3.1.1. Systematic Mapping

This category aims to identify the research space of the literature in the area of testing ML applications. The research questions in this category are:

**RQ1: What is the distribution of papers over time?**
This question will help to identify the trends of publications over time.

**RQ2: Which type of contribution is made in the literature?**
This RQ helps to provide a high-level understanding regarding the contribution of the existing research in the area. It allows reviewers to come up with the classification of existing literature on the basis of contribution to understand the nature of research in the area. The guidelines of Peterson et al. [7] classifies different types of research contribution such as technique, tool, framework or metric, which allows us to assess whether the research community as a whole is focusing towards proposing new techniques or the development of new tools.

**RQ3: Which type of research is presented in the literature?**

This RQ aims to assess the type of research approaches used in the existing papers. Peterson et al. [7] provides the classification of different research approaches, e.g., solution proposal, validation research and evaluation research. This classification helps to determine the maturity of the area in using empirical approaches by slicing the existing literature on the basis of research type. The categories are:

- **Solution Proposal:** A novel solution was proposed for a particular problem and its applicability was evaluated on a small case study or a small example.
- **Validation Research:** A novel technique was proposed and validated in a lab setting through an experiment.
- **Evaluation Research:** A novel technique was evaluated comprehensively through extensive experiments.

**RQ4: Which type of approaches are used in testing of machine learning systems?**
The most fundamental part of any SM is to identify themes and map the existing research in a way to identify key trends and gaps in the area [7]. Therefore, this RQ aims to classify the existing literature by identifying different themes or patterns in the existing literature on testing ML programs. To identify themes, we have used thematic analysis and followed the guidelines given in [18]. According to the guidelines, thematic analysis is performed in six phases. In the first phase, all the authors thoroughly reviewed the existing papers to familiarize themselves with the proposed techniques. In the second phase, the first two authors came up with the initial list of interesting themes in the existing techniques. In the third phase, all these initial themes are gathered and all duplicate themes are removed. In fourth phase, the last two authors analyzed and discussed to merge and refine these themes. In the fifth phase, the naming and definition of these themes were finalized by all the authors. In the last phase, all the included papers were mapped on the defined themes.

**RQ5: Which type machine learning systems are targeted by the testing techniques?**
The type of machine learning system shall be used as an indicator whether the proposed technique aims to test supervised, unsupervised, semi-supervised or reinforcement learning programs. This will help researchers and practitioners to identify the type of learning systems targeted by the existing testing techniques and the type of learning systems that needs more attention in the future.

**RQ6: Which type of testing is proposed for testing of machine learning systems?**
This RQ aims to provide evidence for whether the technique proposed in the paper is black box or white box. Both types of testing are commonly used by the testing community for conventional (non-ML program) testing. Therefore, this question will help researchers and practitioners to know that which types of techniques are proposed in the literature for testing of ML applications.

**RQ7: Which type of test artefact is generated by the proposed techniques?**
This RQ will help to identify the test artefact (e.g. test case, test inputs) generated by the proposed techniques. This help the practitioners of testing ML applications to easily select the technique based on their needs.

**RQ8: What is the subject of study (classifier or ANN) in the proposed technique?**
The objective of this RQ is to assess whether the proposed testing technique targets the classifier (e.g., SVM) or Artificial Neural Network. This will help researchers and practitioners in the identification of suitable technique. Such breakdown of the testing techniques will help to identify gaps regarding the subject which needs more focus from the research community.

**RQ9: Which kind of testing techniques are proposed for ML systems?**
The use of ML is now gaining some traction in mission and safety critical systems (e.g. [19, 20]). It is important to test such systems for both functional and non-functional quality characteristics for example performance, robustness, security and reliability. Therefore, this RQ aims to examine the kind of testing proposed for ML programs. We investigate whether the existing literature contains any non-functional testing techniques (e.g., security) for ML programs or not.

**RQ10: What are the existing testing tools and are they available for researchers and practitioners?**
This question aims to report the tools and their availability for testing ML applications. Identification of available tools helps practitioners in the selection of appropriate tool for their applications.

### 3.1.2. Empirical Evaluation
This category aims to synthesize evidence regarding the empirical evaluations conducted in the area. Empirical evaluations are essential in evaluating the effectiveness of testing techniques and should be articulated properly to ensure their replication as well as for demonstrating the strength and limitations of techniques. The following are the research questions in this category.

**RQ11: Which type of test objects are used in the empirical evaluations?**
This RQ analyze which type of Test Objects (TO) are used in the empirical evaluation. This evidence will help researchers and practitioners regarding the nature of test objects and the applicability and scalability of the proposed testing technique.

**RQ12: What are the evaluation metrics in the existing empirical evaluations?**
This RQ investigate the evaluation metric of the proposed technique that whether the proposed technique is assessed for showing the applicability or other metrics like performance and cost are also taken in consideration by empirical evaluations.

**RQ13: Which type of datasets are used in the evaluations?**
This RQ explore the datasets given as an input to the test object.

### 3.1.3. Quality Assessment
Quality assessment of primary studies is an important component of systematic reviews [9, 21]. It is one of the reliable methods of increasing the level of confidence in the findings of the paper [21]. Therefore, we define the following research questions to assess the quality of papers reported in the area.

**RQ14: What is the reporting quality of the existing papers?**
This research question will help to analyze the quality of published papers on the basis of our five question criteria (given in section 3.4).

### 3.2. Search process
Systematic search process is a key activity of systematic mapping to identify relevant papers [22]. We conducted both manual and automated search for identification of primary papers.

Manual search is conducted in the related important venues and an automated search is conducted by searching in main digital libraries.

### 3.2.1. Manual search

Manual search is conducted by manually searching the most popular venues and journals of software testing listed in Table 2. Based on the title we identified 56 related papers. Additionally, we also searched relevant papers on the personal webpages of the prominent researchers in the area that resulted in 18 related papers. In total, we extracted 74 studies from our manual search process.

Table 2. Popular Conferences, Workshops and journals related to software engineering

| Source | Acronym | Type |
|---|---|---|
| International Conference on Software Testing | ICST | Conference |
| International Conference on Software Engineering | ICSE | Conference |
| International Symposium on Software Testing and Analysis | ISSTA | Conference |
| Automated Software Engineering | ASE | Conference |
| International Workshop on Metamorphic Testing | MET | Workshop |
| Software Testing Verification and Reliability | STVR | Journal |
| Journal of Systems and Software | JSS | Journal |
| Journal of Information and Software Technology | IST | Journal |
| IEEE Transactions on Software Engineering | TSE | Journal |
| IEEE Transactions on Reliability Engineering | TSR | Journal |

### 3.2.2. Automatic search

We conducted automatic search in six widely used [23] digital libraries (listed in Table 3) to gather relevant papers. In order to formulate our search string, we used SEOBook keyword density analyzer[1] to discover the most frequent, 2-word phrases and 3-word phrases in titles, abstract and keywords of papers collected manually. Based on the results of the keywords extracted through SEOBook, we composed the following search string:

(("Software Testing" OR "Metamorphic Testing" OR "Quality Assurance") AND ("machine learning" OR "supervised learning" OR "deep learning" OR "deep neural network" OR "machine learning classifier"))

Table 3. List of digital libraries used in automated search

| Source | URL |
|---|---|
| Google Scholar | https://scholar.google.com.pk/ |
| IEEE Xplore | http://ieeexplore.ieee.org/ |
| ACM Digital Library | http://dl.acm.org/ |
| Springer Link | http://link.springer.com/ |
| Wiley Online Library | http://onlinelibrary.wiley.com/ |
| Science Direct | http://www.sciencedirect.com/ |

The query is run on the above six major digital libraries with full text and specific advance search configuration for each individual library. It is noticeable that the search was completed on 31 January, 2019 and therefore papers published after this date are not part of this paper.

Automatic search resulted in 1579 papers. Therefore, the total papers gathered from both (manual and automatic) our search methods are 1654 which were used for further analysis. Figure 2 illustrates our complete search process.

### 3.3. Study selection

We initially obtained 1654 papers through our manual and automatic search. We define the following inclusion and exclusion criteria for thorough selection of related papers.

*Inclusion criteria*
- Papers that proposed technique, method, tool, framework or an experiment on testing of ML applications.
- Papers which are available online
- Peer-reviewed papers
- Papers written in English
- Papers available in full text
- Papers with multiple versions, only the most recent version is selected

*Exclusion criteria*
- Off the topic papers
- PhD dissertation, white papers, and technical reports that are not published in peer-reviewed venues
- Duplicate papers
- Presentations, magazine articles, book chapters, invited papers, tutorials, lecture notes, monographs, editorials and other non-peer reviewed articles
- Papers not written in English
- Papers not available in full text

At first step, duplicate studies (i.e., a paper present in more than one database) were removed from the initial pool 1654 studies, resulting in the removal of 448 duplicate studies. In the next step, grey literature (presentations, magazine articles, tutorials, lecture notes, editorials, and other non-peer reviewed articles) was removed, which resulted in 706 remaining studies. Irrelevant literature was removed by reading the title and abstract of the studies, which resulted in the final set of 36 studies. To reduce the risk of missing any relevant work, the last two authors of this paper performed snowballing by following the guidelines given by Wohlin et al. [10]. Snowballing is the process of identifying relevant papers from the references (backward snowballing) and citations (forward snowballing) of the selected papers. The newly identified studies are then filtered based on our inclusion and exclusion criteria. In backward snowballing, we examined the references of the selected studies whereas in forward snowballing, we used Google Scholar to analyze the citations of the papers and search for relevant papers which are present in our selected pool of papers. The process of snowballing resulted in the addition of one more paper resulting in a total of 37 studies for further analysis. Figure 2 illustrates the complete process of study selection.

To reduce the bias in the selection of studies, initially the first two authors of this study performed the selection independently and the results of both the authors were then matched. All the inconsistencies between the authors in the selection of studies were discussed and resolved in follow up meetings by the other authors.

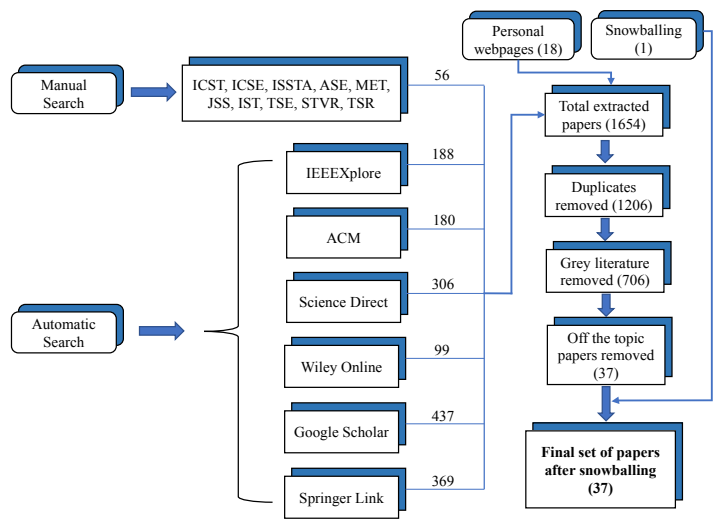

Figure 2. An Illustration of the search process

### 3.4. Quality assessment

Quality assessment is well established activity in systematic reviews [23-25] that helps researchers and practitioners to gain confidence in the results of the paper and conclusions drawn from it [21]. Typically, quality assessment in systematic reviews is performed for selection of papers, difference in quality, weighting each primary paper, interpretation of results or recommendation for future [21]. In this paper, we performed quality assessment for weighting each individual study and recommendations for future research. To thoroughly assess the quality, we adopted a set of five most asked quality assessment questions from [21] to evaluate each included paper on the basis of reporting, rigor, credibility and relevance. Table 4 listed the set of five questions where each question is assigned possible answers of 'Yes', 'Partial' or 'No'. These three possible answers are scored as 1, 0.5, and 0 respectively. Consequently, the quality score for each included paper is computed as the sum of all the scores of the quality assessment questions. We systematically identified quality score for each individual paper. Each paper is evaluated by at least two authors of this study. In case of conflicts regarding the assignment of a quality score between the two authors, a discussion was held among all the authors to reach consensus in several review meetings. It is noticeable that we did not include or exclude papers on the basis of quality. Furthermore, similar criteria for quality assessment is used in several existing systematic reviews (e.g. [21, 26]).

Table 4. List of quality assessment questions

| S.no | Quality Assessment Questions |
|------|------------------------------|
| QC1  | Are the aims of the research articulated? |
| QC2  | Is the proposed technique clearly described? |
| QC3  | Is the experimental design appropriate? |
| QC4  | Is there a clear statement of finding and relate to the aim of the research? |
| QC5  | Does the research add value to academia or industry? |

### 3.5. Data extraction strategy

The final set of 37 papers are taken for detailed analysis to answer our research questions. First, we identified attributes corresponding to each research question which were to be extracted from the included papers. Table 5 provides the detailed listing of these attributes. The first column shows the unique identifiers for attributes listed in the second column. Third column shows the corresponding RQ whereas the last column shows the possible values extracted from papers for each attribute.

We carefully extracted the data against the defined attributes in the spreadsheet by using Google Sheets. The first two authors of this study independently extracted data in separate sheets. Then the results of both the authors were compared to reduce the bias in data extraction. The discrepancies were resolved through multiple review meetings in which both the authors presented the reasons for assigning the possible value. The arguments were judged by the rest of the authors with the aim of reaching consensus.

The spreadsheet is available online (https://bit.ly/2Q4WPGO) for all the researchers and practitioners working in this area. We aim to update the repository at least once a year by adding relevant papers in the future. The summarized results and charts are also present in our repository.

Table 5. List of data attributes with the corresponding RQs and possible values

| ID | Attribute | Corresponding RQ | Possible values |
|---|---|---|---|
| D1 | Year | RQ1 | • Year of publication |
| D2 | Contribution facet | RQ2 | • Technique<br>• Framework<br>• Tool<br>• Experience report |
| D3 | Research facet | RQ3 | • Solution proposal<br>• Validation research<br>• Evaluation research<br>• Others |
| D4 | Testing approach | RQ4 | • Metamorphic testing<br>• Mutation testing<br>• Combinatorial testing<br>• Concolic testing<br>• Multi-implementation testing<br>• AI based approach |
| D5 | Type of machine learning | RQ5 | • Supervised<br>• Unsupervised<br>• Reinforcement |
| D6 | Type of testing | RQ6 | • Black box<br>• White box |
| D7 | Type of artefact | RQ7 | • Test case<br>• Test data |
| D8 | Subject of study | RQ8 | • Classifier<br>• Neural Network |
| D9 | Kind of testing | RQ9 | • Functional<br>• Non-functional |
| D10 | Tool | RQ10 | • Tool name and availability |
| D11 | SUT | RQ11 | • Name of SUT e.g. SVM |
| D12 | Evaluation metric | RQ12 | • Performance<br>• Reliability<br>• Effectiveness |
| D13 | Data Set | RQ13 | • Name of Dataset e.g. MNIST |
| D14 | QC1 | RQ14 | • 0 |

|     |     |      | • 0.5 |
|     |     |      | • 1   |
| D15 | QC2 | RQ14 | • 0   |
|     |     |      | • 0.5 |
|     |     |      | • 1   |
| D16 | QC3 | RQ14 | • 0   |
|     |     |      | • 0.5 |
|     |     |      | • 1   |
| D17 | QC4 | RQ14 | • 0   |
|     |     |      | • 0.5 |
|     |     |      | • 1   |
| D18 | QC5 | RQ14 | • 0   |
|     |     |      | • 0.5 |
|     |     |      | • 1   |

## 4. Results & Discussion

In this section, we discuss the results of the study and presents the synthesis of the data extracted in the previous section to answer the research questions in detail.

### 4.1. RQ1: What is the distribution of papers over time?

The testing of ML programs is a growing area of research. Figure 3 shows the distribution of 37 publications included in this SM from 2007 to 2019. Overall, it seems a dramatic increase in the number of publications from 2015 to 2018. It is noticeable that the search process is conducted until 31 January 2019. Therefore, it might be possible that papers published in the year 2018 and in the first month of 2019 which are not yet indexed are not included in this study. The first seminal work in this area was published in 2007 by Christian Murphy, Gail E. Kaiser, and Marta Arias. The work is cited in subsequent 10 (27%) of our primary studies. The peak year of publications is 2018 in which 20 papers were published in different venues.

Figure 4 shows that conferences are the main venue types with 75% of publications, followed by workshops (18%) and journals (6%). In general, the area is growing with time with an increase in the number of publications every year.

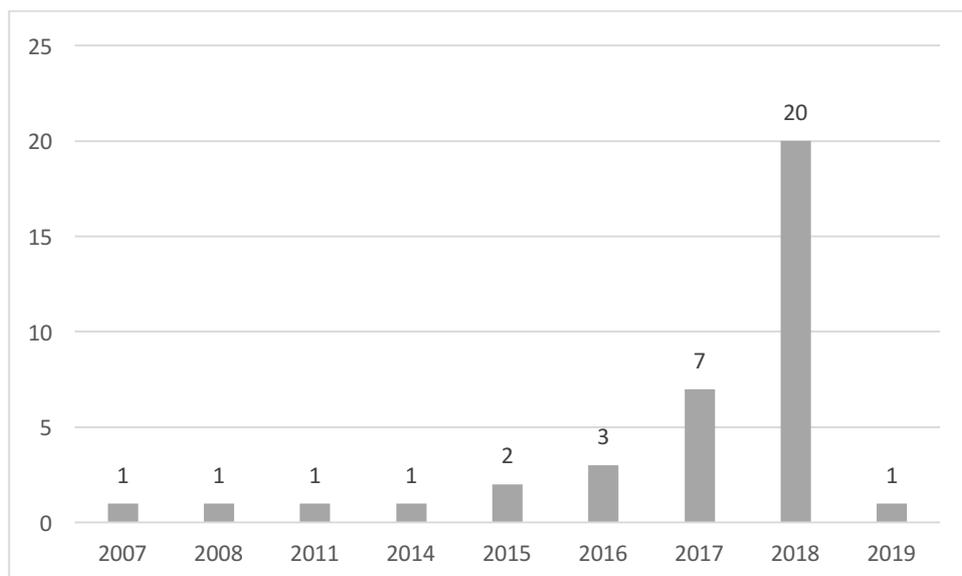

Figure 3. Distribution of papers over time

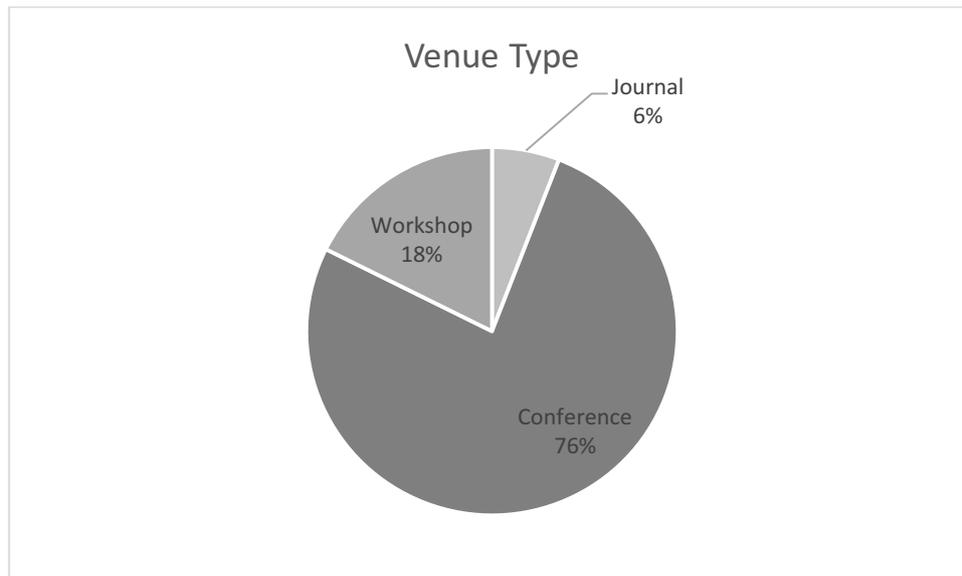

Figure 4. An overview of publications w.r.t venue types

### 4.2. RQ2: Which type of contribution is made in the literature?

We identified different contribution facets in the area of Testing ML programs, as shown in Figure 5. It can be seen from the figure that the major contribution in the area is towards proposing new techniques. Out of 37 papers, 26 presented new testing techniques which indicates an area that is still growing and that does not yet have well established testing approaches. For example, S3 presented a metamorphic testing technique for testing ML classifiers. The technique proposed a set of metamorphic relations which are applied to the target classifier and if any classifier fails to satisfy the relation, it exposes a fault. Nine papers, out of 26, implemented the proposed technique in a tool and are categorized under tool. For example, S23 presented a mutation testing tool and a technique for testing ML programs called DeepMutation. Six papers presented the frameworks out of which three papers implemented the framework in a tool. For example, S13 presented a white box framework based on neuron coverage and a tool called DeepXplore for testing ML programs. Three papers contributed test metrics for ML programs, i.e., S28 proposed different dependability metrics for neural networks and S29 presented K-projection coverage as a metric for testing ML enabled autonomous programs. Only one paper (S11) proposed a method for reducing the need of oracle for testing ML programs and is categorized under 'method'. Consequently, two papers (S6, S24) are categorized under 'others' because they do not present any technique, tool, framework, metric or method but presented a comparative analysis or an empirical evaluation. It is important to note that ten papers presented more than one contribution facet. Out of these ten papers, five presented technique and tool (S7, S18, S22, S23 and S26), three papers (S13, S14 and S21) presented framework and tool, one paper (S4) presented technique and metric and another paper S27 presented technique, metric and tool.

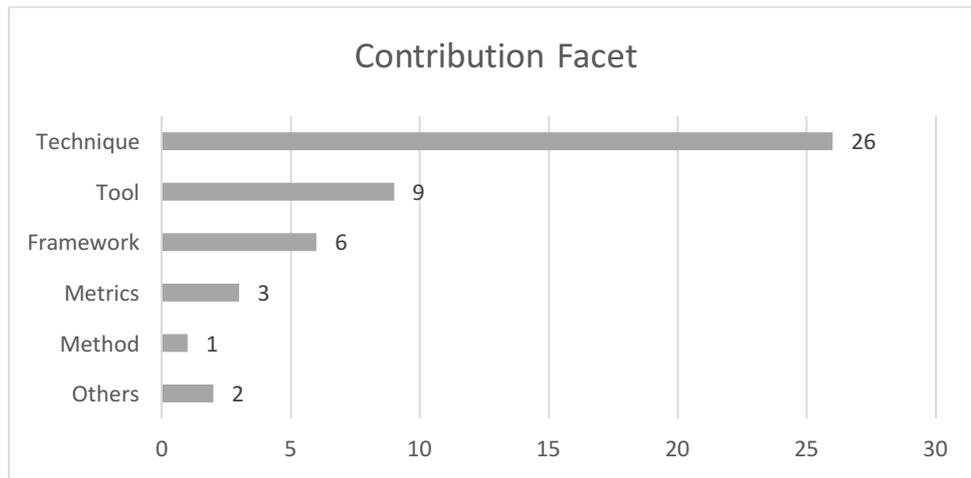

Figure 5. An overview of the types of contribution facet

*4.3.    RQ3: Which type of research is conducted in the literature?*

We identified four research facets in the existing literature of testing ML systems, as shown in Figure 6. It is clearly evident that the research is dominated by validation research and solution proposal in the area. Out of 37 papers, 15 are categorized as validation research as the techniques in these papers are validated in a lab setting through an experiment. For example, S14 evaluated the proposed framework in a lab setting with multiple datasets. Consequently, 15 papers are grouped under the solution proposal as they evaluated the proposed technique with a small example or program. For example, S8 demonstrated the proposed technique on the implementation of a Support Vector Machine algorithm as an example. Only, four papers are grouped as evaluation research because they presented an extensive evaluation of the proposed technique. For example, the technique in S3 is empirically evaluated on the Weka implementation of Naïve Bayes Classifier (NBC) and K-Nearest Neighbors (kNN) as examples. Also, the performance of the proposed technique is compared with cross-validation (which is a common practice in data science) by using mutation analysis. Therefore, S3 is grouped under evaluation research. There are only two papers, S6 and S24 that presented comparative performance analysis and empirical analysis respectively and are categorized as 'Others'. It is noticeable that there is a growing trend in solution proposal and validation research. However, there is a clear lack of research presenting detailed empirical evaluations and comparisons in the area. Therefore, more research is required to comprehensively assess and compare the existing techniques of testing for ML programs.

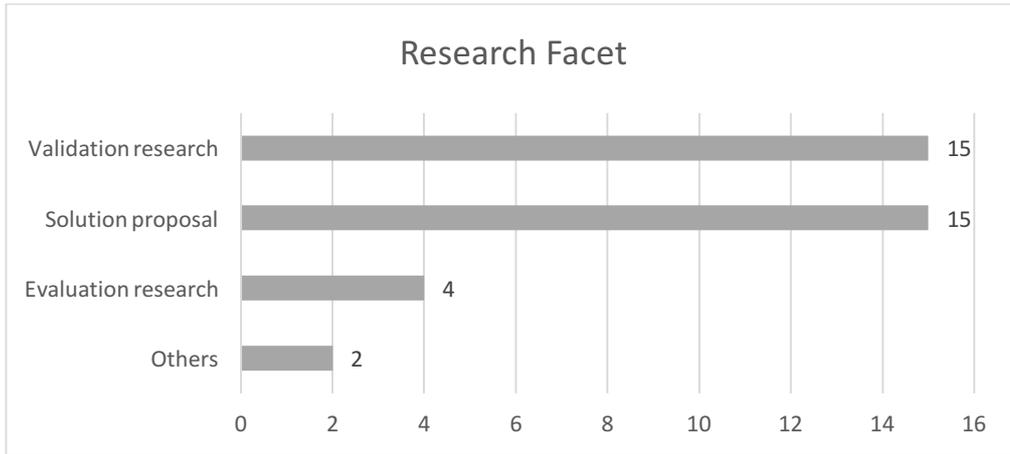

Figure 6. An overview of the types of Research facet

*4.4. RQ4: Which type of approaches are used in testing of machine learning systems?*

We have identified several approaches used by the existing techniques in the literature to test ML programs, as shown in Figure 7. The themes in the figure are identified in a systematic manner (as discussed in section 3.3.1). However, it is important to note that these themes are subjective and based on the existing papers included in this paper. It can be extended or improved as more studies are published in this area.

Our results show that metamorphic testing is the most frequently used approach followed by coverage and adversarial based approaches. Figure 8 provides an overview of the number of papers proposed in each theme. In the following, we discuss each approach in detail.

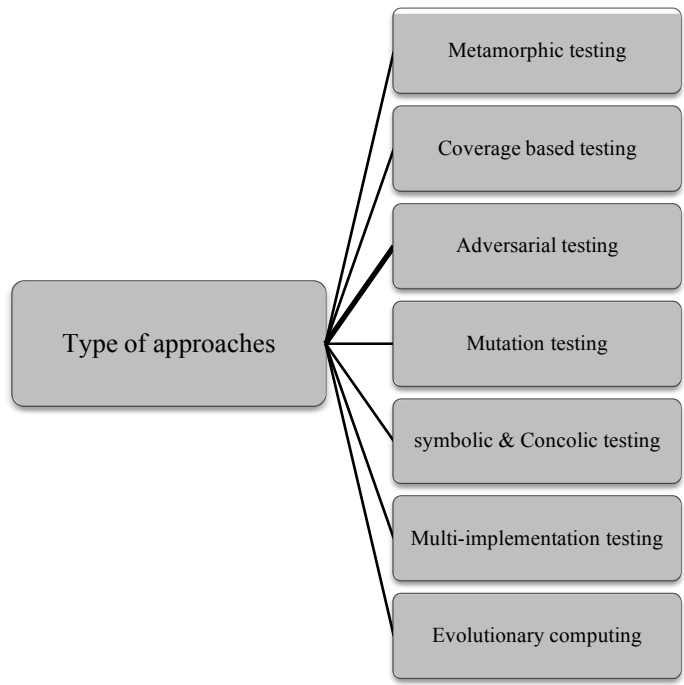

Figure 7. Classification of approaches used to test ML programs

### 4.4.1. Metamorphic Testing (MT)

Out of 35 ML testing techniques, 11 uses the approach of metamorphic testing. The idea of MT was first introduced to ML in S1 by Murphy and his colleagues in 2008. MT is a property-based testing in which metamorphic relations are identified based on the properties of the system. If the system does not hold these relations than it is an indication of a fault in the system. However, these relations are not easy to derive for a particular system and require in-depth knowledge of the domain. Several works are published in the literature to reduce the complexity of deriving metamorphic relations (e.g. [27, 28]). Chen et al. [16] conducted a comprehensive survey on the challenges and opportunities in MT and provides a detailed discussion on metamorphic relations. Another survey is conducted by Segura et al. [29] on the techniques of MT and argued that metamorphic relations can be grouped into six classes, i.e., additive, multiplicative, permutative, invertive, inclusive and exclusive relations. In this paper, we specifically discuss MT approaches proposed for ML programs.

S3 applied MT to supervised ML classifiers and argued that metamorphic relations can represent both necessary and expected properties of a classifier under test. The violation of necessary properties indicates faults in the classifier and serve the purpose of verification. Consequently, the violation of expected properties indicates the divergence between the actual results of the classifier and user expectation, thus, serve the purpose of validation. S16 used feature selection techniques with MT to validate ML classifiers and conducted an empirical evaluation by taking NB and kNN as a case study. S11 discussed the concept of oracle for ML programs by taking SVM as an example. S8 proposed a coverage criterion to systematically derive metamorphic relations for ML classifiers and provide guidelines to perform MT. S10 extended the MT to neural network and presented a framework to validate the classification accuracy of convolutional neural network. S18 proposed a technique and tool called DeepTest to automatically generate test inputs for DNN-enabled autonomous cars by using metamorphic relations for image transformations. S17 extended the use of metamorphic relations to enhance the classification accuracy of classifiers and discover the classification problems. S19 implemented metamorphic testing to discover bugs in the implementation of image classifiers.

In summary, MT seems to be a promising technique to test ML programs. However, there is little effort made for its automation. More research work (like [30]) is needed to provide techniques for automatic detection of metamorphic relations in ML programs. The existing research on testing ML programs is focused towards the verification and validation of classification accuracy. In future, MT can be extended to test other attributes of ML programs such as security, performance and robustness. Moreover, the existing MT techniques targets classifiers and very few are dedicated towards neural networks. Hence, more research work is required to extend the existing MT to other neural networks, e.g., recurrent neural network and deep convolutional network.

### 4.4.2. Coverage Based Testing

Coverage based testing is used in both, traditional and ML programs, as a basis for measuring the quality of software and as an indicator to show confidence in the readiness of software. Recently, several coverage criteria are proposed to test ML programs. S13 proposed neuron coverage for measuring the coverage of Deep Learning (DL) systems by the given test inputs. Neuron coverage measure the number of neurons activated by the test data in a DL model. However, S34 presented neuron combination coverage and argued that simple neuron coverage cannot accurately explore the behavior of a program. Neuron combination coverage measures the number of different combinations of neuron activations in a DL model. S8 introduces a dataset coverage criterion for improving the quality of data distribution in the training data.

S26 discussed combinatorial testing criteria for ML classifiers. Consequently, S27 proposed k-projection coverage for assessing the quality of data in ML programs. S28 discussed different dependability metrics to measure the important attributes of a neural network like correctness, robustness and completeness.

In summary, coverage based approaches in the context of ML programs are still in its infancy and several other criteria are expected to be seen in the future. The above coverage criteria help to measure the internal structure of ML programs for a given inputs or the dataset provided to train ML programs. However, there is lack of research to evaluate and compare the effectiveness and efficiency of these coverage criteria and how they can be used for other testing activities such as test data generation and selection.

### 4.4.3. Adversarial Testing

ML programs have shown good performance in classification tasks in the recent years [31, 32]. However, they are found to be highly unstable to adversarial examples (adding noise/perturbations to the data). Different approaches are proposed in the literature for testing and securing ML programs from adversarial attacks. S7 proposed a DeepFool technique for misleading a deep neural network with high accuracy. S14 proposed a verification framework for feed forward multi-layer deep neural network to explore if it is vulnerable to adversarial attacks. S29 presented a feature guided approach to test the safety of ML classifiers. Consequently, S30 proposed a differential fuzzing approach for creation of adversarial examples by mutating the data with small perturbations and increasing the neuron coverage.

### 4.4.4. Mutation Testing

Mutation testing is the process of evaluating the quality of the existing test cases by modifying the original program with small syntactic changes [33]. It involves the generating mutants by using mutation operators (e.g., arithmetic operator) and executing them against the available test cases. Mutation testing of traditional programs is widely explored in the academia and is a growing area of research [34]. S23 extended the mutation testing to DL programs by proposing a framework called DeepMutation. The framework includes two types of mutation operators i.e., source level mutation operators and model level mutation operators, as shown in Table 6. S36 proposed five mutation operators for two types of operations namely change and deletion in neural networks. S31 conducted an exploratory study on the manifestation of bugs in ML classifiers by using the traditional mutation operators.

We found only three studies related to mutation testing in the context of ML programs and the topic is still understudied. For example, it is argued in the existing studies that insert operation in neural network is more complex, therefore, require more research work [S36]. Moreover, mutation testing has an inherent problem of equivalent mutants and computational cost of the mutants execution [33]. However, the existing studies failed to discuss these limitations and their remedies in the context of testing ML programs, hence needs more research work. In future, more studies are needed to comprehensively evaluate the effectiveness of mutation analysis and proposed more advance mutation operators to cover diverse aspects of ML programs (for example, addition of neuron to a layer or changing the activation function in a layer). Moreover, mutation analysis can be extended to provide automated techniques for detection of bugs and its repair in ML programs [S23].

Table 6. Mutation operators for DL programs [S23]

| Source-level mutation testing operators for DL systems |
|---|

| Fault type | Level | Target | Description |
|---|---|---|---|
| Data Repetition (DR) | Global | Data | Duplicates training data |
|  | Local |  | Duplicates specific type of data |
| Label Error (LE) | Global | Data | Falsify results (e.g., labels) of data |
|  | Local |  | Falsify specific results of data |
| Data Missing (DM) | Global | Data | Remove selected data |
|  | Local |  | Remove specific types of data |
| Data Shuffle (DF) | Global | Data | Shuffle selected training data |
|  | Local |  | Shuffle specific types of data |
| Noise Perturb. (NP) | Global | Data | Add noise to training data |
|  | Local |  | Add noise to specific type of data |
| Layer Removal (LR) | Global | Program | Remove a layer |
| Act. Fun. Remov. (AFRs) | Global | Program | Remove activation functions |
| Layer Addition (LAs) | Global | Program | Add a layer Act. |
| **Model-level mutation testing operators for DL systems** | | | |
| **Mutation operator** | **Level** | **Description** | |
| Gaussian Fuzzing (GF) | Weight | Fuzz weight by Gaussian Distribution | |
| Neuron Effect Block. (NEB) | Neuron | Block a neuron effect on following layers | |
| Weight Shuffling (WS) | Neuron | Shuffle selected weights | |
| Neuron Activation Inverse (NAI) | Neuron | Invert the activation status of a neuron | |
| Neuron Switch (NS) | Neuron | Switch two neurons of the same layer | |
| Layer Addition (LAm) | Layer | Add a layer in neuron network | |
| Layer Deactivation (LD) | Layer | Deactivate the effects of a layer | |
| Act. Fun. Remov. (AFRm) | Layer | Remove activation functions | |

### 4.4.5. Symbolic or Concolic Testing

Symbolic testing is used to determine inputs to execute different paths in a program [35]. Similarly, concolic testing is the blend of symbolic analysis and program execution used to generate test inputs that provide better coverage [36]. Recently, S22 extended concolic testing to Deep Neural Networks (DNNs) to increase the requirement coverage. The approach first select's inputs in the test suite that is close to satisfying a requirement and uses symbolic execution to obtain a new test input that satisfy the requirement. S33 and S35 provide a technique for security analysis of DNNs based on symbolic interval analysis.

Symbolic or concolic testing in the context of ML programs are understudied and more approaches can be seen in future to scale these approaches for large DNNs and other ML classifiers.

### 4.4.6. Evolutionary computing

Three papers are categorized under this theme because they use an evolutionary approach for testing ML programs by using evolutionary algorithms. S5 generates images by using MAP-Elites evolutionary algorithm that can easily fool a neural network. The approach is evaluated on two popular datasets called ImageNet and MNIST. S25 uses a multi-objective search and decision tree classification model to test vision-based control systems. The search is being guided by the classification model towards critical scenarios. S34 presented a framework called Telemade that used an evolutionary approach for generating adversarial images. The framework consists of three main components, i.e., test input generation to generate adversarial images, test input validation to verify the test inputs and a coverage criterion called neuron combination coverage to measure the effectiveness of the generated test inputs.

### 4.4.7. Multi-implementation testing

Multi-implementations approaches were traditionally recommended for critical systems [37-39]. However, such approaches are recently proposed for ML programs to overcome the oracle problem because these approaches do not require an explicit oracle when used for the purpose of testing. Primarily, S20 introduced the concept of multi-implementation testing to ML classifiers. However, the main limitation of these approaches is the cost and availability of multiple implementations of the same program. To overcome this limitation, S26 proposed an approach called SynEva to automatically generate a mirror program by using program synthesis.

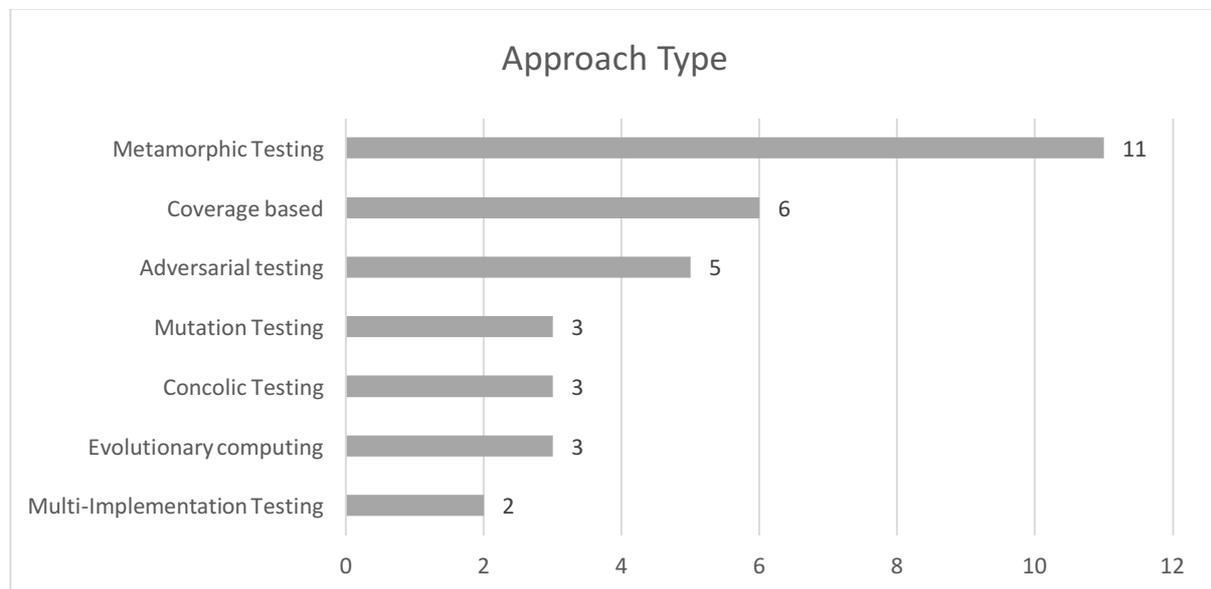

Figure 8. An overview of the types of approaches used for testing ML systems

Above we discuss different types of approaches proposed in the literature to test ML programs. The approaches are grouped together in different themes depending on the nature of the proposed approach. Only one paper, S34, is classified under two themes, coverage based testing and evolutionary computing because it proposed a coverage criterion as well as an evolutionary approach for generating test inputs. In future, more hybrid approaches can be seen that combines more than one approach and used as complementary with each other. For example, mutation testing can be used with coverage based approaches to strengthened its effectiveness. Similarly, MT testing can be improved when used in combination with a coverage criterion (e.g. neuron coverage).

### 4.5. RQ5: Which type of testing is proposed for testing of machine learning systems?

About 74% (26) of the papers proposed black box approaches and 26% proposed white box approaches to test ML programs, as shown in Figure 9. For example, S29 provides a feature-guided black box approach to test the safety of a neural network irrespective of knowing its internal details. Nine papers (26%) proposed white box testing techniques to test ML programs.

For example, S13 proposed a white box testing framework called DeepXplore to test neural network by directly interacting with its internal components (neurons). In future, more techniques may be researched in particular white box techniques for testing ML programs as very little work is currently available. Moreover, there is a need for comparing and assessing both types of approaches to assess whether they outperform each other and do they complement each other in the detection of bugs or not?

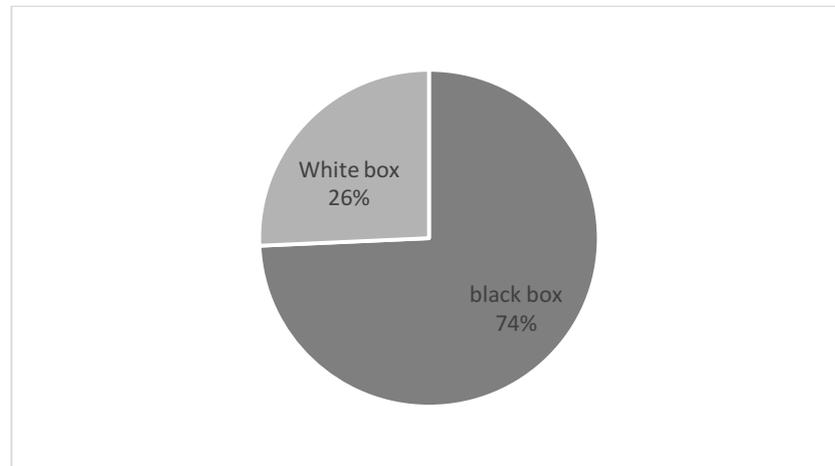

Figure 9. Type of testing used to test ML programs

*4.6.    RQ6: Which type of machine learning systems are targeted by the testing techniques?*
The existing techniques for testing targeted two types of ML programs namely supervised and unsupervised learning programs, as shown in Figure 10. It is interesting to note that about 17 papers (about 49%) proposed generic techniques that work for testing of both supervised and unsupervised learning programs. For example, S4 proposed metamorphic relations which are applicable for both supervised and unsupervised learning programs. Similarly, S12 applied a combinatorial testing technique that works for supervised and unsupervised learning ML classifiers. In particular, 16 papers targeted only supervised learning programs to test. For example, S3 provides the technique of metamorphic testing to test supervised ML classifiers. Similarly, S19 proposed a testing technique based on metamorphic testing for supervised ML programs, i.e., classifiers and deep neural network. In general, 65% (34) of the techniques support the testing of supervised learning and 33% (16) support the testing unsupervised learning programs.

In the future, more attention is required to study the applicability or extension of the existing testing techniques for other types of learning programs such as semi-supervised learning and reinforcement learning. More new techniques are expected in the future to overcome the limitations of the existing techniques. For example, devising more rigorous metamorphic relations for testing ML programs.

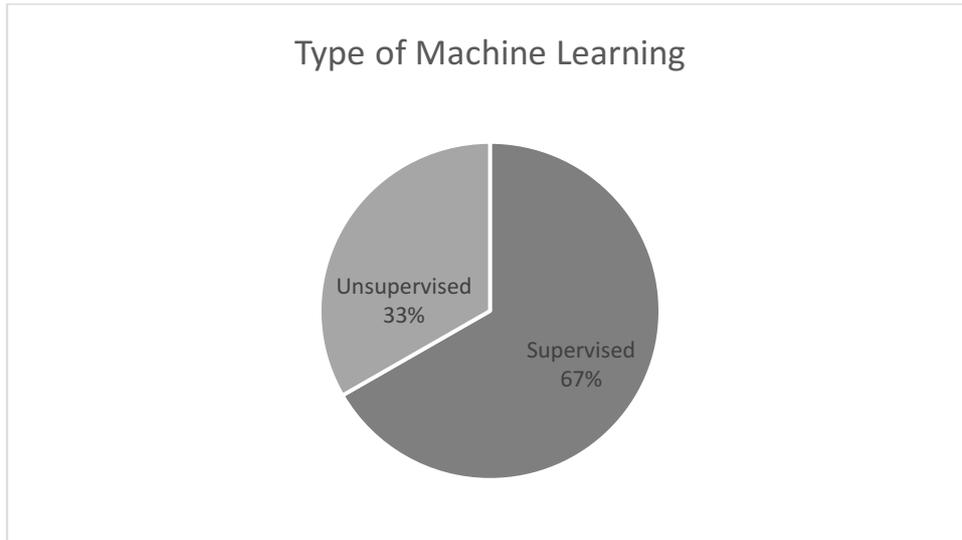

Figure 10. Type of machine learning targeted by the proposed testing approaches

*4.7. RQ7: Which type of test artefact is generated by the proposed techniques?*
An overview of the generated test artifacts is shown in Figure 11. Most of the papers (22) created test cases in their proposed techniques. For example, S3 generated metamorphic test cases by proposing various metamorphic relations for ML classifiers. About 27% (10) papers generated test data to test ML programs. For example, S5 and S7 generated image data to fool neural networks and identified bugs. Six papers provide test requirements for ML programs. Test requirements are not actual test inputs or test cases but the condition that can be used to generate test cases or test inputs. For example, S13 proposed a coverage criterion called neuron coverage which can be used as a condition to generate test data for neural networks. It is notable that seven papers created more than one artifact in their techniques. For example, S3 created test cases and test data whereas S34 created test data and provide test requirement (coverage criterion). Two papers are categorized as 'Others' because they provide empirical evaluation rather than generating any test artifact.

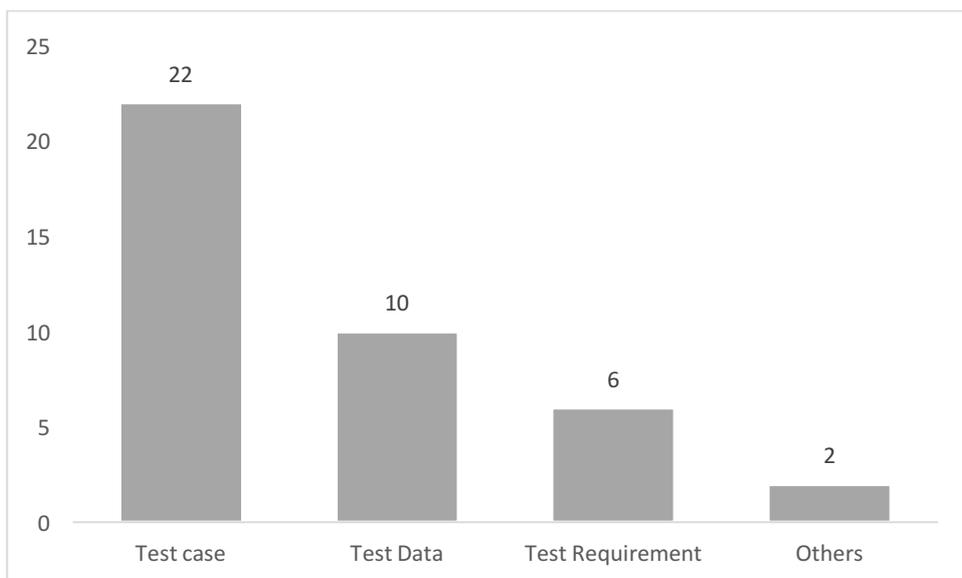

Figure 11. An overview of the generated test artefact

*4.8. RQ8: What is the subject of study (classifier or ANN) in the proposed technique?*

In general, the existing testing techniques in this area aims to test either ML classifiers or neural networks, as shown in Figure 12. About 57% (21) papers proposed techniques to test ANN or ANN enabled programs, e.g., S13. Similarly, about 43% (16) papers presented techniques to test ML classifiers, e.g., S2. It is noticeable that four papers (S10, S11, S28, S34) proposed generic techniques that are applicable to test both, i.e., ANN and classifiers.

Overall, metamorphic testing seems a dominant technique to test ML programs (classifiers and neural networks). However, more research work is required to construct new significant metamorphic relations. The existing techniques of metamorphic testing in this area are based on the properties of ML program whereas data is also significant component in such applications. In future, more metamorphic relations can be constructed based on the properties of data to verify the correctness of training data for ML programs.

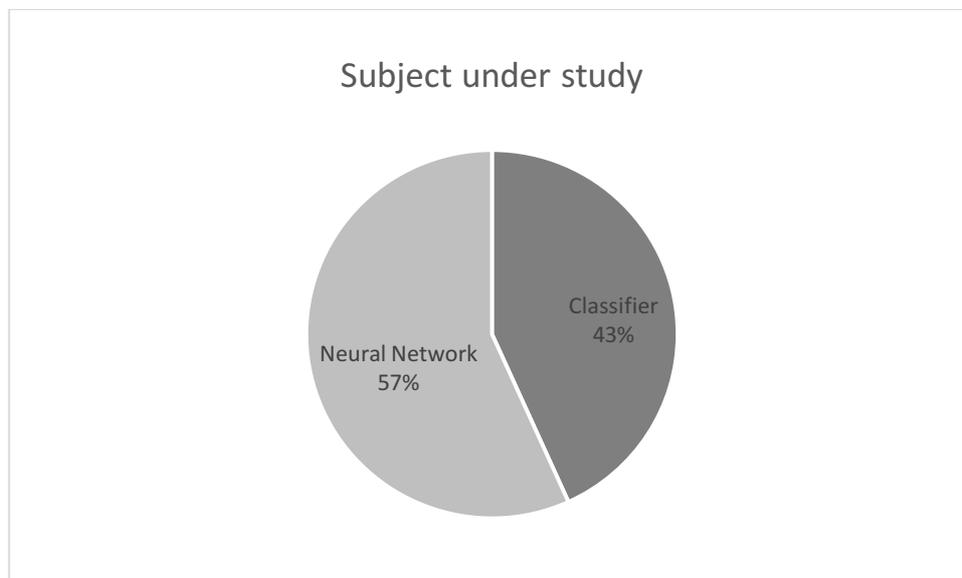

Figure 12. An overview of the subject under study

*4.9. RQ9: Which kind of testing techniques are proposed for ML systems?*

As anticipated, most of the proposed techniques (30 out of 35) in the literature focused on the functional testing (classification accuracy and generality) of ML programs. However, we found five papers (S14, S29, S33, S34, S37) that focused on security and safety analysis of ML programs. For example, S29 proposed a feature-guided black box technique for safety verification of deep neural networks. Similarly, S37 proposed a technique for security analysis of deep neural network to verify its vulnerability to adversarial examples. Figure 13 shows an overview of the kind of techniques proposed in the existing literature.

In summary, ML programs are normally tested for its accuracy and generality. However, the assessment of other quality attributes in neural network requires more research.

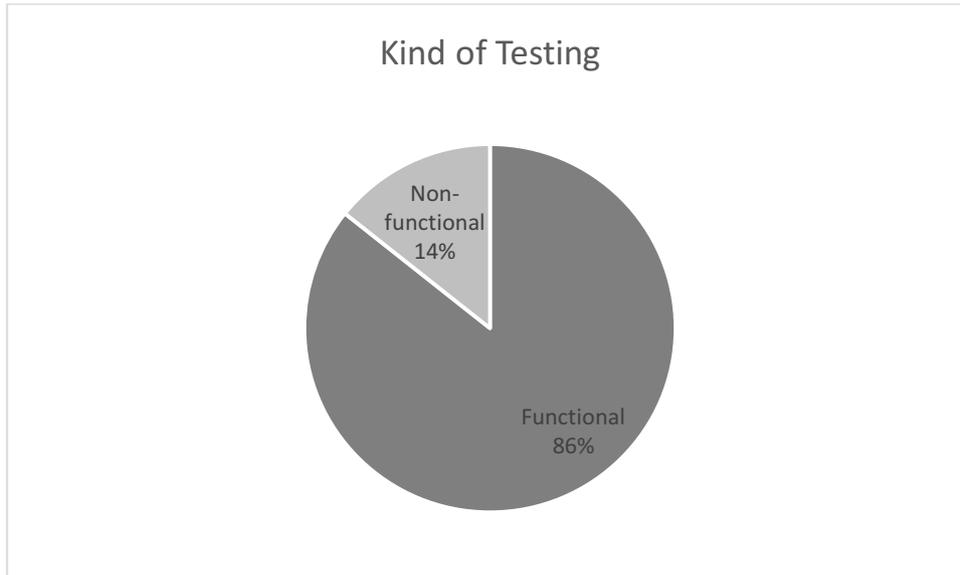

Figure 13. An overview of the kind of testing

*4.10. RQ11: What are the existing tools and their availability for researchers and practitioners?*

We identified several tools proposed in the existing papers, listed in Table 7. We believe it is a positive sign for practitioners and researchers that tools are being developed in this area for practical use of the proposed technique. However, only nine papers out of 37 presented tools that can be used by practitioners. But, upon further investigation regarding the availability of the tools we found that only three tools (DeepFool, DeepXplore and DeepConcolic) are available for download. The non-availability of tools can limit the use of the proposed technique for practitioners and other researchers.

Table 7. List of tools proposed in the papers

| Tool | Availability | Reference |
|---|---|---|
| DeepFool | Yes | S7 |
| DeepXplore | Yes | S13 |
| Deep Learning Verification (DLV) | No | S14 |
| DeepTest | No | S18 |
| DeepRoad | No | S21 |
| DeepConcolic | Yes | S22 |
| DeepMutation | No | S23 |
| SynEva | No | S26 |
| Telemade | No | S34 |

*4.11. RQ12: Which type of SUT's are used in the empirical evaluations?*

We identified several SUTs in the included papers, shown in Figure 14. Overall, the empirical evaluations conducted in this area are on trivial ML classifiers or algorithm except S25 which conducted the evaluation on an industrial case study. This shows that the area is still in its infancy and require more research to gain maturity. It can be seen that most of the papers (19 out of 27) used to build and train their neural network (NN) for assessing the effectiveness of the proposed testing technique. However, only a few papers reported the details of the constructed artificial neural network and these models itself can be susceptible to multiple biases. Therefore, there is a need for public repositories of models that can be used by the testing community to assess the effectiveness and efficiency of their proposed techniques.

Support Vector Machine (SVM) is the most used (10 out of 27) classifier used in the papers followed by K-Nearest Neighbor (5 out of 27) and Naïve Bayes (5 out of 27). Two papers use C4 algorithm and another two use MArtiRank algorithm in their empirical assessment. Algorithms such as Random Forest, Priori, Expectation-Maximization (EM), K Means, Decision Tree and Logistic Regression are used only once in a paper as a test objects. One of the paper (S2) applied the proposed technique to an anomaly-based intrusion detection system called PAYL [40].

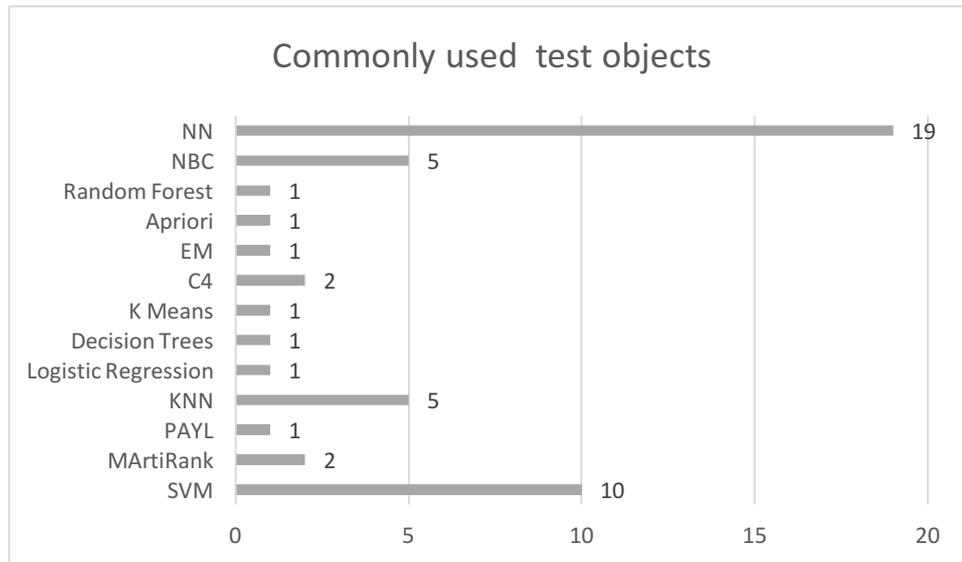

Figure 14. An overview of the System Under Tests (SUTs)

Figure 15 shows the number of classifiers used in each paper. Only ten papers evaluated their technique on two to five test objects. The rest of the papers uses only one test object to assess the effectiveness of the proposed technique. This raises concern about the generality and scalability of the proposed techniques. Therefore, more empirical research needs to be conducted to thoroughly assess the effectiveness of the proposed techniques on large test objects.

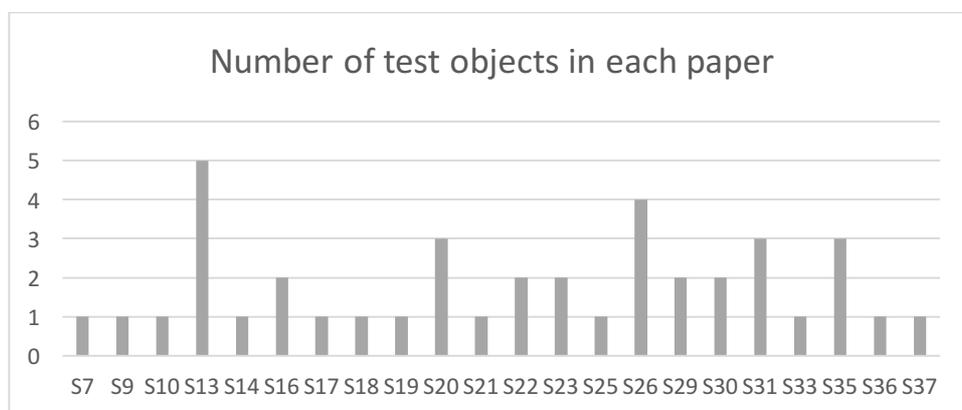

Figure 15. Number of test objects used in each paper

## 4.12. RQ13: What are the evaluation metrics in the existing empirical evaluations?

About 76% (27 out of 35) papers show the applicability of the proposed by applying it to the target test object, expecting to detect the real bugs. However, only 12% (4 out of 35) of the papers shows the effectiveness of the technique by using mutation analysis along with the applicability. For example, S3 measured the effectiveness of the proposed technique by using mutation analysis. Consequently, only 12% (4 out of 35) of the papers measure the performance (computational cost) of the proposed technique. For example, S22 assess the efficiency of its proposed technique by comparing them with other state-of-the-art techniques. Figure 16 provides an overview of the evaluation metrics.

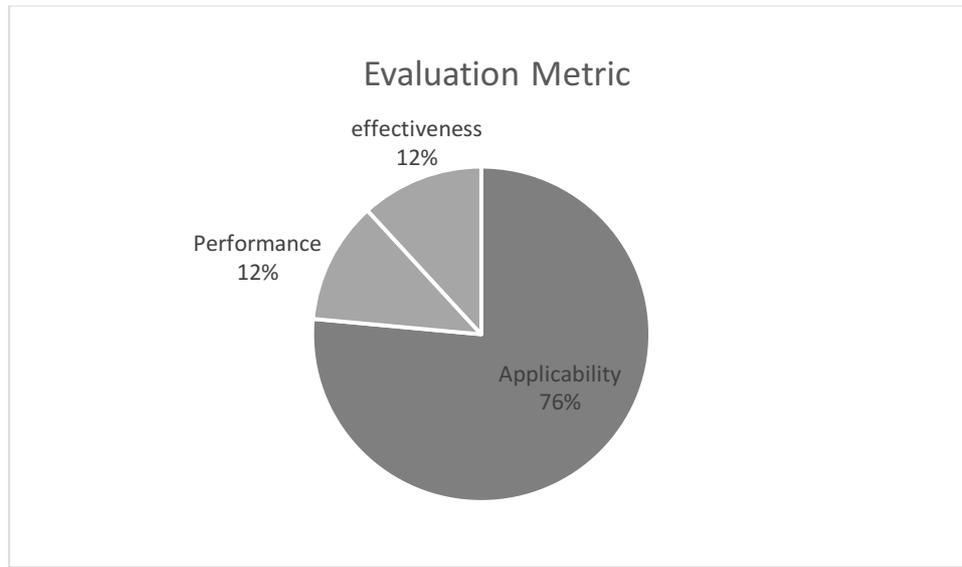

Figure 16. Evaluation metrics of the proposed techniques

## 4.13. RQ14: Which type of datasets are used in the evaluations?

Several datasets are used in the existing literature during the empirical investigation, as shown in Table 8. It can be seen that MNIST is the most used dataset followed by Udacity car and CIFAR. Out of 35 studies that reported empirical investigation, 11 uses MNIST dataset of handwritten digits. The dataset contains the training set of 60,000 examples and a test set of 10,000 examples. Consequently, four studies used Udacity Car that contains driving data for self-driving cars and another four studies used CIFAR dataset which contains 60,000 images for 10 different objects. ImageNet, Drebin, IRIS and IRIS waveform are used in at least two studies. Moreover, the detailed list of datasets used at least once in the included studies is given in the table.

Table 8. List of datasets used in the empirical evaluations

| ID | Dataset Name | Frequency | References |
|---|---|---|---|
| 1 | MNIST | 11 | S5, S7, S13, S19, S22, S23, S29, S30, S33, S35, S36 |
| 2 | Udacity Car Dataset | 4 | S13, S18, S21, S35 |
| 3 | CIFAR | 4 | S7, S22, S23, S29 |
| 4 | ImageNet | 2 | S5, S30 |
| 5 | Drebin | 2 | S13, S35 |
| 6 | IRIS | 2 | S20, S26 |
| 7 | IRIS waveform | 2 | S26, S31 |

| 8 | German | 1 | S31 |
|---|---|---|---|
| 9 | kr-vs-kp | 1 | S31 |
| 10 | Balance scale weight and distance | 1 | S26 |
| 11 | DSRC vehicle communication | 1 | S26 |
| 12 | Breast Cancer | 1 | S20 |
| 13 | Glass identification | 1 | S20 |
| 14 | Artcodes | 1 | S17 |
| 15 | ILPD | 1 | S16 |
| 16 | Vertebral column data | 1 | S16 |
| 17 | p-DI | 1 | S10 |
| 18 | ISLVRC | 1 | S7 |
| 19 | Bag of words | 1 | S9 |
| 20 | VirusTotal | 1 | S13 |
| 21 | KTTI | 1 | S37 |

*4.14. RQ15: What is the reporting quality of the existing papers?*

Overall, the reporting quality of the papers can be ranked as good because more than 90% of the papers (34 out of 37) have a quality score greater than or equal to 3.0 (out of 5.0). Figure 17 shows the quality of each included paper on the basis of our five-score quality criteria. The average quality of all the included studies is 4.01 (out of 5.0) which means that most of the papers are of good quality based on our quality criteria. However, it can be seen from Figure 18 that most of the papers (30 out of 37) fail to provide appropriate experimental design (QC3) for evaluation of the proposed technique. Eleven studies failed partially (score 0.5) or completely (score 0) to describe the proposed technique clearly.

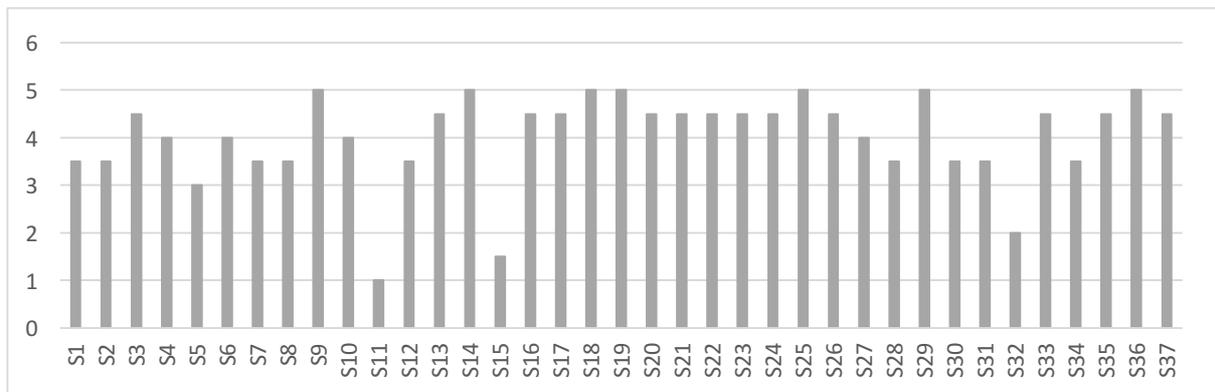

Figure 17. Reporting quality of each included paper

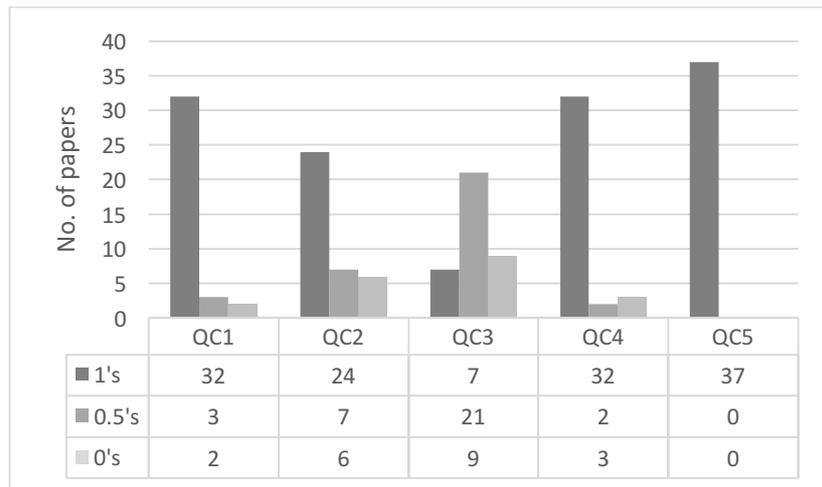

Figure 18. The number of 1's, 0.5's and 0's in each Quality Criterion

Table 9 provide a detailed listing of each quality criterion and the corresponding score for each included study.

Table 9. A detailed listing of the quality assessment

| ID  | QC1 | QC2 | QC3 | QC4 | QC5 | SUM |
|-----|-----|-----|-----|-----|-----|-----|
| S1  | 1   | 0.5 | 0   | 1   | 1   | 3.5 |
| S2  | 1   | 0.5 | 0   | 1   | 1   | 3.5 |
| S3  | 1   | 1   | 0.5 | 1   | 1   | 4.5 |
| S4  | 1   | 0.5 | 0.5 | 1   | 1   | 4   |
| S5  | 0.5 | 0.5 | 0   | 1   | 1   | 3   |
| S6  | 0.5 | 1   | 0.5 | 1   | 1   | 4   |
| S7  | 1   | 0.5 | 0.5 | 0.5 | 1   | 3.5 |
| S8  | 1   | 0.5 | 0   | 1   | 1   | 3.5 |
| S9  | 1   | 1   | 1   | 1   | 1   | 5   |
| S10 | 1   | 1   | 0   | 1   | 1   | 4   |
| S11 | 0   | 0   | 0   | 0   | 1   | 1   |
| S12 | 1   | 0   | 0.5 | 1   | 1   | 3.5 |
| S13 | 1   | 1   | 0.5 | 1   | 1   | 4.5 |
| S14 | 1   | 1   | 1   | 1   | 1   | 5   |
| S15 | 0   | 0   | 0   | 0.5 | 1   | 1.5 |
| S16 | 1   | 1   | 0.5 | 1   | 1   | 4.5 |
| S17 | 1   | 1   | 0.5 | 1   | 1   | 4.5 |
| S18 | 1   | 1   | 1   | 1   | 1   | 5   |
| S19 | 1   | 1   | 1   | 1   | 1   | 5   |
| S20 | 1   | 1   | 0.5 | 1   | 1   | 4.5 |
| S21 | 1   | 1   | 0.5 | 1   | 1   | 4.5 |
| S22 | 1   | 1   | 0.5 | 1   | 1   | 4.5 |
| S23 | 1   | 1   | 0.5 | 1   | 1   | 4.5 |
| S24 | 1   | 1   | 0.5 | 1   | 1   | 4.5 |
| S25 | 1   | 1   | 1   | 1   | 1   | 5   |
| S26 | 1   | 1   | 0.5 | 1   | 1   | 4.5 |
| S27 | 1   | 0.5 | 0.5 | 1   | 1   | 4   |
| S28 | 0.5 | 1   | 0   | 1   | 1   | 3.5 |
| S29 | 1 | 1 | 1 | 1 | 1 | 5 |
| S30 | 1 | 1 | 0.5 | 0 | 1 | 3.5 |
| S31 | 1 | 0 | 0.5 | 1 | 1 | 3.5 |
| S32 | 1 | 0 | 0 | 0 | 1 | 2 |
| S33 | 1 | 1 | 0.5 | 1 | 1 | 4.5 |
| S34 | 1 | 0 | 0.5 | 1 | 1 | 3.5 |
| S35 | 1 | 1 | 0.5 | 1 | 1 | 4.5 |
| S36 | 1 | 1 | 1 | 1 | 1 | 5 |
| S37 | 1 | 1 | 0.5 | 1 | 1 | 4.5 |

**5. Threats to Validity**

This study has several threats like other systematic mapping studies. However, we have taken several measures to validate and mitigate their effects. In this section, we discuss various threats to validity related to our search, selection, data extraction, and quality assessment of papers.

5.1. Search

One possible threat can be the inappropriate selection of search terms and missing of relevant literature. We presented detailed discussion in section 3.2.2 regarding the formulation of query and search process. In order to reduce this threat, we used two search strategies i.e. manual and automated. The papers selected through manual search were used direct our further search process. We used an automated tool for the identification of appropriate search terms to formulate the query for an automated search. Additionally, we performed the process of snowballing, forward and backward snowballing, to reduce the risk of missing relevant literature. Further to reduce the risk, we have also searched personal webpages of the prominent researchers in the area.

5.2. Selection

In order to reduce bias in the selection of papers, we let the first two authors to independently perform the selection by using our inclusion and exclusion criteria. The results of both the authors were matched. There were only two disagreements which were resolved by involving other authors.

5.3. Data extraction

To reduce the bias in the data extraction, two authors independently extracted data from the selected papers. Each paper was carefully analyzed w.r.t to the identified attributes and the data was extracted in an online spreadsheet shared with all the authors. The results of both the authors were verified and matched by the other two authors of this study. All the discrepancies regarding the classification were resolved in several review meetings. In case of any missing information, an email was sent to the corresponding author of that study for the required information to ensure the complete and correct data extraction.

5.4. Quality assessment:

To reduce the bias in assessing the reporting quality of the papers, we selected the top five most used quality assessment questions [21] and the quality of each study was assessed on the basis of those five questions. In response to each question, each study was assigned the value of 1, 0.5 or 0. The assignment of values for each study was conducted in the same fashion like data extraction and the quality of each paper was verified by at least two authors.

## 6. Conclusion

The goal of the study is to gather, analyze and classify the current state of the art in the area of testing ML programs which will help practitioners and researchers in several ways. It provides an overview of the state-of-the-art in the area and can be used as a catalog of the existing testing tools and techniques used to test ML programs. Our results show a significant increase in the number of publications in recent years with the most number of papers (76%) published in conferences as compare to journals (6%) and workshops (18%). Several different tools and techniques are proposed based on different approaches. However, only few tools are publically available and also there is lack of enough empirical evidence to assess and compare the existing techniques. We present a detail taxonomy of the available techniques that identifies several gaps and set the ground for future research in this area. The taxonomy highlights popular themes or approaches in the area such as metamorphic testing, coverage based testing, adversarial testing, mutation testing, symbolic and concolic testing, multi-implementation testing and evolutionary computing. Each approach entails its own limitations, which needs to be alleviated and is a subject for future research in this area. For example, metamorphic testing is based on metamorphic relations which are highly domain specific and are hard to construct. Similarly, mutation testing comes with the limitations of equivalent mutants and computational cost which are yet to be studied in the context of ML programs. Also, additional mutation operators can be proposed to increase the presentation of faults in ML programs.

We found that most of the proposed techniques (74%) are black box, which means these techniques do not need internal details (i.e., code) of a test object. In contrast, one-fourth (about 26%) of the proposed techniques are white box that requires the internal details of the test object. Upon further investigation, we discover that about 67% of the testing techniques target ML programs with supervised learning and about one third (33%) of the techniques target programs with unsupervised learning. At present, we found no technique in the existing literature that discusses the testing of ML programs build with semi-supervised or reinforcement learning approaches. More than half (57%) of the papers proposed techniques to test neural networks while less than half (43%) of the papers focus on ML classifiers. Subsequently, about 85% of the techniques are classified as functional testing techniques as they test the accuracy of ML programs. However, only 14% of the techniques test the performance, security and safety in ML programs and are classified as non-functional testing techniques.

We identify different test objects used in the empirical assessments. It is found that most the papers that targeted ML classifiers have used SVM followed by k nearest neighbor (kNN) and Naïve Bayes Classifier (NBC). On contrary, 19 papers who have targeted neural network in their testing techniques have developed their own neural networks reporting only few details. Additionally, we identify the metrics evaluated in the empirical evaluations regarding the proposed technique and the datasets used. It is evident that most of the papers (76%) evaluated only the applicability of the proposed technique. However, about 12% of the papers used mutation analysis to evaluate the effectiveness of the proposed technique and another 12% evaluated the performance of the proposed technique as well. Also, the most used dataset is MNIST being used in 11 papers followed by Udacity Car Dataset and CIFAR with the frequency of four papers each.

Finally, we assessed the reporting quality of the included studies by using a standard five questions quality criteria. Overall, the quality of the papers can be ranked as good because majority of the papers have quality score equal to or greater than 3.0. However, only few papers successfully meet QC3 which indicates that most of the papers failed to report enough details regarding the experiment and poses threat to the its repeatability.

In future there is a need to focus on evaluation of the approaches in real industrial contexts. Most of the presented works are content to demonstrate the applicability of their approach without thorough comparison with existing approaches. There is significant room for research on testing ML programs that use semi-supervised and reinforcement learning. There are few tools available to practitioners and it is important that more tools are developed and made available to the wider research and practitioner community.

Table 10. List of all included papers

| ID | Author | Title | Year |
| --- | --- | --- | --- |
| S1 | Murphy *et al.* [41] | An Approach to Software Testing of Machine Learning Applications | 2007 |
| S2 | Murphy *et al.* [42] | Properties of Machine Learning Applications for Use in Metamorphic Testing | 2008 |
| S3 | Xie *et al.* [43] | Testing and Validating machine learning classifiers by metamorphic testing | 2011 |
| S4 | Groce *et al.* [44] | You Are the Only Possible Oracle: Effective Test Selection for End Users of Interactive Machine Learning Systems | 2014 |
| S5 | Nguyen *et al.* [45] | Deep Neural Networks are Easily Fooled: High Confidence Predictions for Unrecognizable Images | 2015 |
| S6 | Aleem *et al.* [46] | Comparative Perfomance Analysis of Machine Learning Techniques for Software Bug Detection | 2015 |
| S7 | Dezfooli *et al.* [47] | DeepFool: a simple and accurate method to fool deep neural networks | 2016 |
| S8 | Nakajima *et al.* [48] | Dataset Coverage for Testing Machine Learning Computer Programs | 2016 |
| S9 | Ribeiro *et al.* [49] | Why Should I Trust You?" Explaining the Predictions of Any Classifier | 2016 |
| S10 | Ding et. [50] | Validating a Deep Learning Framework by Metamorphic Testing | 2017 |
| S11 | Nkajima *et al.* [51] | Generalized Oracle for Testing Machine Learning Computer Programs | 2017 |
| S12 | Chandrasekaran *et al.* [52] | Applying Combinatorial Testing to Data Mining Algorithms | 2017 |
| S13 | Pei *et al.* [53] | DeepXplore: Automated Whitebox Testing of Deep Learning Systems | 2017 |
| S14 | Huang *et al.* [54] | Safety verification of deep neural networks | 2017 |
| S15 | Sellam *et al.* [55] | I Like the Way You Think!" Inspecting the Internal Logic of Recurrent Neural Networks | 2017 |
| S16 | Azni *et al.* [56] | Validation of Machine Learning Classifiers Using Metamorphic Testing and Feature Selection Techniques | 2017 |
| S17 | Xu *et al.* [57] | Enhancing Supervised Classifications with Metamorphic Relations | 2018 |
| S18 | Tian *et al.* [58] | DeepTest: automated testing of deep-neural-network-driven autonomous cars | 2018 |
| S19 | Dwarakanath *et al.* [59] | Identifying implementation bugs in machine learning based image classifiers using metamorphic testing | 2018 |
| S20 | Srisakaokul *et al.* [60] | Multiple-Implementation Testing of Supervised Learning Software | 2018 |
| S21 | Zhang *et al.* [61] | DeepRoad: GAN-Based Metamorphic Testing and Input Validation Framework for Autonomous Driving Systems | 2018 |
| S22 | Sun *et al.* [62] | Concolic Testing for Deep Neural Networks | 2018 |
| S23 | Ma *et al.* [63] | DeepMutation: Mutation Testing of Deep Learning Systems | 2018 |
| S24 | Zhang *et al.* [64] | An Empirical Study on TensorFlow Program Bugs | 2018 |
| S25 | Abdessalem *et al.* [65] | Testing Vision-Based Control Systems Using Learnable Evolutionary Algorithms | 2018 |
| S26 | Qin *et al.* [66] | SynEva: Evaluating ML Programs by Mirror Program Synthesis | 2018 |
| S27 | Cheng *et al.* [67] | Quantitative Projection Coverage for Testing ML-enabled Autonomous Systems | 2018 |
| S28 | Cheng *et al.* [68] | Towards Dependability Metrics for Neural Networks | 2018 |

| S29 | Wicker et al. [69] | Feature-guided black-box safety testing of deep neural networks | 2018 |
| S30 | Guo et al. [70] | DLFuzz: Differential Fuzzing Testing of Deep Learning Systems | 2018 |
| S31 | Cheng et al. [71] | Manifesting Bugs in Machine Learning Code: An Explorative Study with Mutation Testing | 2018 |
| S32 | Nishi et al. [72] | A Test Architecture for Machine Learning Product | 2018 |
| S33 | Wang et al. [73] | Formal Security Analysis of Neural Networks using Symbolic Intervals | 2018 |
| S34 | Yang et al. [74] | Telemade: A Testing Framework for Learning-Based Malware Detection Systems | 2018 |
| S35 | Wang et al. [75] | Efficient Formal Safety Analysis of Neural Networks | 2018 |
| S36 | Shen et al. [76] | MuNN: Mutation Analysis of Neural Networks | 2018 |
| S37 | Tuncali et al. [77] | Simulation-based Adversarial Test Generation for Autonomous Vehicles with Machine Learning Components | 2019 |